# A modified single and multi-objective bacteria foraging optimization for the solution of quadratic assignment problem

Saeid Parvandeh*, Pariya Soltani, Mohammadreza Boroumand, Fahimeh Boroumand

Department of Molecular and Human Genetics, Baylor College of Medicine, Houston, TX, U.S.
Tandy School of Computer Science, University of Tulsa, Tulsa, OK, U.S.
Email: saeid-parvandeh@utulsa.edu, saeid.parvandeh@bcm.edu
* Corresponding author

**Abstract:** Non-polynomial hard (NP-hard) problems are challenging because no polynomial time algorithm has yet been discovered to solve them in polynomial time**.** The Bacteria Foraging Optimization (BFO) algorithm is one of the metaheuristics algorithms that is mostly used for NP-hard problems. BFO is inspired from the behavior of the bacteria foraging such as Escherichia coli (E-coli). The aim of BFO is to eliminate those bacteria that have weak foraging properties and maintain those bacteria that have breakthrough foraging properties toward the optimum. Despite of strength of this algorithm, in most of the problems reaching to optimal solutions are time-demanding or impossible. In this paper, we modified single objective BFO by adding a mutation operator and multi-objective BFO (MOBFO) by adding mutation and crossover from genetic algorithm operators to update the solutions in each generation, and local tabu search algorithm to reach the local optimum solution. Additionally, we used fast nondominated sort algorithm in MOBFO to find the best nondominated solutions in each generation. We evaluated the performance of the proposed algorithms through a number of single and multi-objective Quadratic Assignment Problem (QAP) instances. The experimental results show that our approaches outperform to some previous optimization algorithms in both convergent and divergent solutions.

## 1 Introduction

The goal of optimization process is to find the optimal solution with minimum cost for non-deterministic polynomial hard (NP-hard) problems (Alothaimeen, Ibraheem; Arditi 2019; Gunantara 2018). Basically, there are two types of NP-hard problems, single objective and multi-objectives (DEB 2004). Evolutionary Algorithm (EA) is considered as an effective method to find optimal solutions for such problems (Ojha et al. 2019; Emmerich and Deutz 2018). And, there are many state-of-the-art evolutionary computation algorithms that have been used to solve the combinatorial optimization problems such as Nondominated Sorting Genetic Algorithm II (NSGA-II) (Deb et al. 2002), Strength Pareto Evolutionary Algorithm Version 2 (SPEA2) (Zitzler and Thiele 1999), Multi-Objective Evolutionary Algorithm based on Decomposition MOEA/D (Zhang and Li 2007), and Pareto-Archived Evolution Strategy (PAES) (J. D. Knowles and Corne 2000). Recently, (Li and Wang 2017) improved cuckoo search algorithm by adding combination of the learning-evolve thought with Gaussian distribution to improve the convergence velocity and optimization accuracy in unconstrained function optimization problems. (Wang and Zhang 2016) improved artificial bee colony algorithm by updating a new search equation for onlooker bees, where optimal and suboptimal solutions were considered in iteration process, and an opposition-based method applied to initial and final solution to enhance the global convergence. (Wang, Jie-Sheng; Song 2017) combined chaotic mapping strategy and the biogeography-based optimization optimal migration model and applied on function optimization problems. (Luo, Chuanwen; Yin, Xiaowei; Ni 2015) proposed a discrete bacterial foraging optimization (DBFO) based on the idea of BFO algorithm. They applied binary encoding to solve both discrete and continuous optimization problems and rotation step to update velocities of bacteria. Then, they evaluated the efficiency of algorithm through three classical benchmark functions and spectrum allocation problem of cognitive radio. (Yi et al. 2016) applied multi-objective BFO (MOBFO) on aluminum electrolysis production process to find the optimal solutions. They used PAEA approach and Adaptive Foraging Strategy (AFS) to balance convergence and diversity of Pareto front. (F. Zhao et al. 2016) proposed a novel BFO to solve the permutation flow-shop scheduling problem (PFSP), that is a typical combinatorial NP-hard problem with discrete solution space. They showed that convergent speeds and entrapment in the local optimum that were incurred by original BFO can be handled by combining a differential evolution operator and a chaotic search operator in chemotactic part of BFO algorithm. (Darvishi et al. 2014) used MOBFO to solve the optimal power flow (OPF), that is a multi-objective NP-hard problem. They proposed a new BFO algorithm by embedding the fuzzy strategy to original BFO to eliminate the problem of choosing the penalty factors for constraints and behave them just like objective function. Most of the proposed methods modified either chemotactic part or elimination-dispersal part of BFO algorithm for the single or multi-objective problems.

Quadratic Assignment Problem (QAP) and multi-objective QAP (mQAP) are the NP-hard problems so that the solutions for them cannot be found in polynomial time. Thus, there should be an optimization algorithm to solve such NP-hard problems. Recently, many Evolutionary Algorithms (EA) have been used for the solution of QAP and mQAP. (Tosun 2014) used Genetic Algorithm (GA) (Tosun, Dokeroglu, and Cosar 2013) to solve single objective QAP. He proposed a new recombination of operator based on order-1 crossover algorithm and used quick sort algorithm to generate different chromosomes partitions. (Benlic and Hao 2015) proposed a population based memetic algorithm (BMA) for the solution of single objective QAP. They used an effective local optimization algorithm Breakout Local Search (BLS) (Benlic and Hao 2013) within the evolutionary computing framework for improvement of the results. (Zinflou, Gagné, and Gravel 2013) used Genetic Immune Strategy (GISMOO) for the solution of mQAP. GISMOO is hybrid between GA and Artificial Immune System (AIS) to find the solutions in mQAP. The updating part of this algorithm is classical GA operators and AIS cloning selection principle. (Michalak 2016) proposed a new local search method that utilizes the knowledge concerning promising search directions and can be used as a general framework and combined with an EA for the solution of mQAP and Traveling Salesman Problem (TSP). (Shukla 2015) used bat algorithm for the solution of QAP. They suggested that Bat algorithm cannot be directly used for solving discrete search space problem like QAP, thus they used Smallest Position Value (SPV) a heuristic rule to enable the bat algorithm for the solution of QAP.

In this paper, we modified BFO for single objective problems and MOBFO for multi-objective problems. For single objective BFO, we added swap mutation into chemotactic part to avoid the repeated permutations and local tabu search algorithm (James, Rego, and Glover 2009) to the end of the algorithm in order to optimize the cost of final permutation and avoid the local optimum. For MOBFO, we added swap mutation and Uniform-Like Crossover (ULX) operators (Tosun 2014; Tosun, Dokeroglu, and Cosar 2013) in chemotactic part and local tabu search algorithm to the end of the algorithm. Additionally, we used fast nondominated sort method (Deb et al. 2002) into elimination and dispersal part of the MOBFO algorithm to optimize obtained nondominated set solutions in each iteration. We applied modified BFO and MOBFO on QAP and mQAP problems, respectively. Our contributions for this paper are: first, we modified BFO and MOBFO algorithms for the purpose of single and multi-objective problems. Second, regardless of the problem size, these algorithms can find compatible solutions for both QAP and mQAP. Third, however the optimum solutions may not be found eventually, but the experimental results show that the average percentage gap of the solutions for QAP are lower than state-of-the-arts algorithm and the nondominated solutions for mQAP are very close to Pareto front.

## 2 Quadratic Assignment Problem

Quadratic Assignment Problem (QAP) is one of the well-known nonpolynomial hard (NP-hard) problems that is introduced by (Koopmans and Beckmann 1957). It can be defined as a problem to allocate a facility set to a location set with mutual distances among the locations and mutual flows among the facilities. Given two $n \times n$ matrices $A = (f_{ij})$ and $B = (d_{kl})$ as inputs with real valued elements, and let $f_{ij}$ be flow between facility $i$ and facility $j$, and let $d_{kl}$ be the distance between the location $k$ and location $l$. Also, let $n$ be the number of facilities and locations, where $n = (1, 2, ..., n)$ QAP is a single objective optimization problem with one objective function and can be defined as:

$$\min_{\varphi \in S_n} Q(\varphi) = \sum_{i=1}^{n} \sum_{j=1}^{n} f_{ij} d_{\varphi(i)\varphi(j)} \tag{1}$$

where, $S_n$ is a permutation set of $n$ locations and $f_{ij}d_{\varphi(i)\varphi(j)}$ is the cost of the assigning facility $i$ to location $\varphi(i)$ and facility $j$ to location $\varphi(j)$. The aim in QAP is to assign each facility to a location with minimum cost (Burkard et al. 1998).

Unlike single objective optimization problems, multi-objective optimization problems involve multiple conflicting objectives that often approach by aggregating the objectives into a scalar function and solves single objective optimization problem. (J. Knowles and Corne 2003) introduced multi-objective QAP (mQAP) with multiple flow matrices and one distance matrix. mQAP is mostly used where one facility should assign to one location with respect to multiple diverse flow matrices and one distance matrix. The mQAP is defined as:

$$\min_{\varphi \in S_n} \overline{Q}(\varphi) = \{Q^1(\varphi), Q^2(\varphi), \ldots, Q^m(\varphi)\} \tag{2}$$

where,

$$Q^p(\varphi) = \sum \sum f_{ij}^p d_{\varphi(i)\varphi(j),} 1 \le p \le m \tag{3}$$

where, $f_{ij}^p$ indicates $p^{th}$ flow between facility $i$ and facility $j$, and $m$ is the number objectives. Other definitions in QAP work here for mQAP. Thus, the result of mQAP is a set of nondominated tradeoff solutions, such that none of the solution dominated by others.

## 3 Bacteria Foraging Optimization

BFO algorithm is one of the bio-inspired optimization algorithms that is inspired from the biomimicry of the E-coli bacteria. BFO introduced by (K M Passino 2002; Kevin M Passino 2010) and its concept is to eliminate the bacteria that have the weak foraging properties and maintain the bacteria that have breakthrough foraging properties toward the nutrient collection or maximize the energy per unit time. Each bacterium communicates with other bacteria by sending the signals, and bacteria move to the next step for collecting nutrient if previous factors have been satisfied. Initially, BFO introduced to make a bridge between microbiology and engineering, and mimics some properties of the bacteria foraging such as chemotactic, reproduction, and quorum sensing, but later in 2010, he investigated and compared BFO with other optimization algorithms (Kevin M Passino 2010). Basically, BFO consists of four principal parts: 1) chemotactic, 2) swarming, 3) reproduction, and 4) elimination and dispersal (Das et al. 2009), and a brief introduction of each part is given as follows:

### *3.1 Chemotactic*

In biology, the chemotactic is a process of bacteria movement for gaining the nutrient. One of the E-coli bacterium properties is that, it can move in two diverse ways, swimming and tumbling. In swimming, the bacterium swims in the same direction to search for nutrient, and in tumbling, bacterium changes the direction to another direction. Assume $\theta^i$ (*j, k, l*) shows the current position of $i^{th}$ bacterium, $j^{th}$ chemotactic step, $k^{th}$ reproduction step, and $l^{th}$ elimination and dispersal step, then the position of bacterium in the next chemotactic step by tumbling will be as:

$$\theta^i(j+1,k,l) = \theta^i(j,k,l) + C(i)\frac{\Delta(i)}{\sqrt{\Delta^T(i)\Delta(i)}} \tag{4}$$

where, $C(i)$ shows the number of steps in the random direction that is specified by the tumbling of $i^{th}$ bacterium and $\Delta(i)$ indicates a vector of random directions in population size with continuous values between [-1, 1].

### *3.2 Swarm*

A group of E-coli bacteria arrange themselves in traveling ring by moving toward the nutrient collection. When the cells stimulated by a high level of succinate, release an attractant aspartate, such that the cells aggregate into groups and move as concentric patterns of swarms with high bacteria density. The cell-to-cell signaling in E-coli swarm can be represented as following function:

$$\begin{aligned} J_{cc}(\theta, P(j,k,l)) &= \sum_{i=1}^{S} J_{cc}\left(\theta, \theta^i(j,k,l)\right) \\ &= \sum_{i=1}^{S}\left[-d_{attract}\exp\left(-w_{attract}\sum_{m=1}^{P}\left(\theta_m - \theta_m^i\right)^2\right)\right] \\ &+ \sum_{i=1}^{S}\left[-h_{repelent}\exp\left(-w_{repelent}\sum_{m=1}^{P}\left(\theta_m - \theta_m^i\right)^2\right)\right] \end{aligned} \tag{5}$$

where, $J_{cc}(\theta ,P(j, k, l))$ shows objective function cost and in each step the amount of this objective function is added to the main cost (where the total cost to be minimized). In other words, $J_{cc}$ shows how far a bacterium is from the fittest bacterium. $\theta$ is the position of $p$-dimensional search space of $i^{th}$ bacterium, $j^{th}$ chemotactic step, $k^{th}$ reproduction step, and $l^{th}$ elimination-dispersal step, $P$ is number of variables to be optimized in each bacterium $i$, $S$ is the total number of bacteria, and $d_{attract}$, $w_{attract}$, $h_{repellent}$, $w_{repellent}$ are the diversity coefficients (constant values) (Das et al. 2009).

### *3.3 Reproduction*

After chemotactic and swarming parts got completed, some of the bacteria have enough nutrient and some others not. In the reproduction part, those bacteria that have enough nutrient will be reproduced and others will be eliminated. To do so, a health status of each bacterium is calculated. The health status is the sum of step fitness during its life and can be defined as:

$$J_{health}^{i} = \sum_{j=1}^{N_c} J(i,j,k,l) \tag{6}$$

where, $N_c$ is total number of chemotactic steps, and $J_{health}$ is calculated for $i^{th}$ bacterium, $j^{th}$ chemotactic step, $k^{th}$ reproduction step, and $l^{th}$ elimination and dispersal step. Eventually, the health status of all bacteria will be sorted in ascending order and the first half of the bacteria reproduce and surrogate into second half of the bacteria based on their top health status. In other words, the smaller $J_{health}$ values related to healthier bacteria. Thus, bacteria with smaller health values have more chance to survive. This process not only keeps the population constant, but also the healthier bacteria continue to next generation.

### *3.4 Elimination and dispersal*

When the density of bacteria getting high in a small area, the temperature of this area getting high and may not be enough nutrient for all bacteria as well. Thus, in this part, the population of bacteria may randomly change their positions to avoid these flaws. Elimination and dispersal event relocates the bacteria in different environments to avoid the bacteria death and local optimum solution(s).

In single objective BFO, the aim is to find one single solution, whereas the aim of the MOBFO is to obtain a set of nondominated solutions to be closed to optimum Pareto front. Thus, (Niu et al. 2013) described MOBFO approach to solve multi-objective optimization problems. The idea of integration between bacteria health sorting and Pareto front mechanism are developed to search for nondominated set solutions of multi-objective problems. Consequently, BFO algorithm can be used for single objective problems and multi-objective problems.

## 4 Proposed Method

QAP is one of discrete optimization problems that can be introduced in two different ways: single objective QAP and multi-objective QAP. First, we describe the proposed single objective BFO algorithm for the solution of single objective

QAP, and we continue with the proposed MOBFO for the solution of mQAP in the next section.

### 4.1 Single Objective BFO

**Algorithm 1**

1. Parameter initialization:
   $p$, $S$, $N_c$, $N_s$, $N_{re}$, $N_{ed}$, $S_r$, $P_{ed}$,
   $p$: dimension of search space
   $S$: population of bacteria,
   $N_c$: chemotactic parts for each bacterium,
   $N_{re}$: reproduction part,
   $N_{ed}$: elimination dispersal part,
   $S_r = S/2$: bacteria split,
   $P_{ed}$: probability of elimination-dispersal.
2. Generate random permutation for each bacterium $i = 1, 2, \ldots, S$, and compute the cost using fitness function $J(i, j, k)$.
3. Sort the costs in ascending order and select the minimum one as best so far.
4. Elimination-dispersal counter: $ell = ell + 1$.
5. Reproduction counter: $k = k + 1$.
6. Chemotactic counter: $j = j + 1$.
   (a) Take the chemotactic part for $i^{th}$ bacterium, $i = 1, 2, \ldots, S$ as follows:
   (b) Apply mutation on each bacterium $i^{th}$.
   (c) Compute the objective function $J(i, j, k)$.
   (d) Sort the costs in ascending order and replace the minimum cost with the best so far if that is smaller.
   (e) Go to (b) if $(i + 1) \neq S$.
   (f) Get the minimum cost so far.
   (g) Go to step 6 if $j + 1 < N_c$.
7. Sort bacteria cost in descending order and remove the second half of the population and duplicate the first half in order to keep the population constant. Go to step 5 if $k + 1 < N_{re}$.
8. Regenerate random permutation for all bacteria based on a random probability value $P_{ed}$.
9. Get minimum cost and replace with the best so far if that is smaller.
10. Apply tabu search algorithm on the permutation related to best so far cost. Compute the cost of new permutation and replace with best so far, if that is smaller. Go to step 4 if $ell + 1 < N_{ed}$.

We proposed a new BFO algorithm by adding swap mutation operator into the chemotactic part and local search algorithm at the end of algorithm to be applied to final solution. Swap mutation (Soni, Nitasha; Kumar 2014) operator is one of the genetic operators that is used to update the solutions in a population. Given a permutation of a QAP solution, swap mutation selects two locations of the solution space and exchanges them, this happens only one time for each solution in the population and the population gets updated for next run. In modified BFO, we generated a random number between [0, 1] for each permutation and if the random number is greater than 0.5, we swapped the entities of two random locations, otherwise we divided the permutation into three blocks and applied the swap mutations in each block. Tabu search algorithm (James, Rego, and Glover 2009) is one of the well-known local search algorithms that searches through solution space locally and periodically to improve the solution $x$ to $x'$ around $x$. Basically, tabu search is mostly used for NP-hard problems and it varies from other local search algorithms based on the tabu list. Tabu list is a typical short-term memory that includes previously visited solutions and stores some of the attributes of the solutions. Thus, it doesn't allow to revisit the solutions so that the algorithm examines those movements that are not in the tabu list. Consequently, the tabu search algorithm is applied on final solution, and the new solution will be replaced with old solution if it produces a better result. The aim in the proposed single objective BFO is to find a single permutation solution with minimum cost in QAP and it is designed in three parts: 1) chemotactic, 2) reproduction, and 3) elimination and dispersal.

The proposed single objective BFO is given in Algorithm 1 and consists of 10 steps. In step 1, all the variables are initialized with preconceived constant numbers (Table 1). In step 2, a set of permutations with $p$ size are randomly generated and fitness function is computed using Equation (1) for each permutation.

In step 3, the minimum cost through all solutions is set as best so far. In step 4, 5, and 6 the chemotactic, reproduction, and elimination and dispersal parts are started, respectively. In step 6, the chemotactic part takes place, where we added swap mutation to update the permutations. We compute the cost using Equation (1) again for all updated permutations, and we substitute the minimum cost if that is less than best so far. In step 7, all permutations are sorted in ascending order based on their cost and the first half of the solutions are substituted into second half. We always use an even number of bacteria (Table 1) to keep the population constant. In step 8, we generate a random number between 0 and 1, and if the generated random number is bigger than $P_{ed}$ (0.25), step 2 will be repeated here, otherwise the algorithm continues to step 9 with previous solutions. In step 9 we find the minimum cost among the permutation costs and if it is smaller than best so far then it will be replaced. In step 10, where we applied tabu search algorithm, the best so far solution gets optimized with tabu search algorithm and substitutes with the old solution if that is better, and it goes to step 4.

**Table 1** Initial settings for algorithm parameters and coefficients

| Variables | Setting |
|---|---|
| Dimension of search space ($p$) | 50 |
| Number of bacteria per generation ($S$) | 100 |
| Chemotactic steps ($N_c$) | 100 |
| Reproduction steps ($N_{re}$) | 4 |
| Elimination-dispersal steps ($N_{ed}$) | 5 |
| Probability for elimination-dispersal ($P_{ed}$) | 0.25 |

### 4.2 Multi-objective BFO

Unlike single objective BFO that is designed for single objective optimization problems, MOBFO is designed for multi-objective optimization problems. The proposed single objective BFO described in previous section and finds a single optimum solution in QAP, whereas the proposed MOBFO finds a nondominated set solutions. In multi-objective optimization problems, solution *a* is better than solution *b* if and only if solution *a* dominates solution *b.* (Deb et al. 2002) introduced fast nondominated sort method to compare different solutions in the multi-objective optimization problems and find the best nondominated set solutions. Briefly, a fast nondominated sort method goes through each solution $i^{th}$ and computes the number of solutions that dominate $i^{th}$ solution and save them in a list. The final list is number of dominated solutions per each solution $i^{th}$. Eventually, the algorithm sorts the list in ascending order and selects those solutions that have zero rank and print them as a nondominated set solutions or Pareto front solutions. We use fast nondominated sort method for the solution of mQAP that is a minimization problem, although there are other nondominated sort algorithms that are used to find the maximized solutions set (Srinivas and Deb 1994; Ghosh and Das 2008; Yazdi, Choi, and Kim 2017).

Like proposed single objective BFO, we added a swap mutation operator into chemotactic part and improved the final solution using local tabu search algorithm. Additionally, we added one crossover operator into chemotactic part in order to update the nondominated set solutions. Crossover is one of the main operators in GA that randomly exchanges some elements in the parent chromosomes and generates the new offspring in the next generation. There are many kinds of crossover methods that can be used to update the population, but a few of them can be used for permutation style population (Misevicius and Kilda 2005).

**Algorithm 2**

1. Parameter initialization:
   $p, S, N_c, N_s, N_{re}, N_{ed}, S_r, P_{ed},$
   $p$: dimension of search space
   $S$: population of bacteria,
   $N_c$: chemotactic parts for each bacterium,
   $N_{re}$: reproduction part,
   $N_{ed}$: elimination dispersal part,
   $S_r = S/2$: bacteria split,
   $P_{ed}$: probability of elimination-dispersal.
   $M$: number of fitness functions.
2. Generate random permutation for each bacterium $i = 1, 2, \ldots, S,$ and compute the cost using fitness function $J_b\ (i, j, k)$, where $b$ is number of fitness functions, $b = 1, 2, \ldots, M$.
3. Get nondominated set solutions using fast nondominated sort algorithm.
4. Elimination-dispersal counter: $ell = ell + 1$.
5. Reproduction counter: $k = k + 1$.
6. Chemotactic counter: $j = j + 1$.
   (a) Take the chemotactic part for $i^{th}$ bacterium, $i = 1, 2, \ldots, S$ as follows:
   (b) Apply crossover and mutation on each bacterium $i^{th}$.
   (c) Compute the fitness function $J_b\ (i, j, k)$.
   (d) Get the nondominated set solutions again.
   (e) Go to (b) if $(i + 1) \neq S$.
   (f) Store all nondominated solutions in each iteration here these are basis for nondominated set solutions.
   (g) Let bacteria with better rank continue for next iteration. Go to step 6 if $j + 1 < N_c$.
7. Sort the solutions in descending order using fast nondominated sort algorithm and remove the second half of the population and duplicate the first half in order to keep the population constant. Go to step 5 if $k + 1 < N_{re}$.
8. Regenerate random permutation for all bacteria based on a random probability value $P_{ed}$.
9. Get nondominated set solutions. Apply multi-objective tabu search algorithm on final nondominated set solutions. Compute the fitness functions of new solutions and replace with old nondominated set solutions, if new ones are better. Go to step 4 if $ell + 1 < N_{ed}$.

Thus, we investigated these kinds of crossover by evaluating the results of each and turns out that Uniform Like Crossover (ULX) (Tate and Smith 1995; Misevicius and Kilda 2005) can be the best one among them.

Briefly, the ULX compares two permutations *A* and *B* and copies the common elements into a new solution (permutation). For the rest of the uncommon elements, it starts with first element of permutation *A* and copies into new permutation, then switch on the permutation *B* and copies the first uncommon element to the next empty location in the new permutation if there is no repeated element in the new permutation, otherwise switches to other uncommon element in another permutation. This process continues for all uncommon elements in the permutations *A* and *B* to fill out the new permutation empty locations. Note, if all of the uncommon elements of both permutations have been visited and still some of the locations remains empty in the new permutation, then a random unduplicated number will be generated to fill out any empty location. In modified MOBFO, after inserting common element of permutation *A* and *B* into a new permutation list, we generated a random number between [0, 1], and if the random number is greater than 0.5, we begin copying of elements from permutation *A*, otherwise we begin with permutation *B*.

The proposed MOBFO is given in Algorithm 2 and consists of 10 steps. The steps of the algorithm are similar to the proposed single objective BFO (Algorithm 1), though at some points it is designed for multi-objective optimization and we describe them here. In step 1, there is an additional variable *M*, that shows the number of objective functions under study. In step 2, a random permutation is created for each bacterium and MOBFO computes *M* objective functions for each bacterium. In step 3, a nondominated set solutions are computed using fast nondominated sort algorithm. In step 6, in addition to swap mutation, ULX is applied to update the solutions and step 3 is repeated. In step 9, step 3 is repeated again to update nondominated set, and tabu search algorithm is applied on the final nondominated set to optimize the solutions locally, and it goes to step 4.

## 5 Experimental Results

We were assessed the performance of the proposed single objective BFO algorithm using the instances from the well-known QAPLIB website (Burkard, Karisch, and Rendl 1997). We ran experimental analyses on a P4 Laptop with 1.0 GB RAM and 2.7 GHz Intel CPU. The operating system was Windows 7 and the developing software was MATLAB 2013a.

Table 2 shows the comparison of the proposed BFO, original BFO, and GA (Tosun 2014) on QAPLIB instances. The comparison is based on the percentage gap between the solution found and optimum solution reported in QAPLIB. In this table, the first column is the name of instances, the second column is the size of instances, third column is the best solution found using the proposed BFO algorithm, and last three columns are the percentage gap of the proposed BFO, original BFO, and GA, respectively. The comparison of the results indicates that the proposed BFO found the optimum solution in most of the instances and outperforms the results of original BFO and GA. These instances selected among more than 100 instances in QAPLIB to test the performance of the proposed algorithm with different problem size (12-256).

**Table 2** List of the name, size, and cost of instances obtained from QAPLIB. Last three columns indicate the comparison of percentage gap between the proposed BFO (Prop. BFO), original BFO (Orig. BFO), and GA algorithms. NA indicates not available.

| Name | Size | Cost | Prop. BFO | Orig. BFO | GA |
|---|---|---|---|---|---|
| chr12a | 12 | 9552 | 0.10 | 0.64 | NA |
| chr15a | 15 | 9896 | 0.12 | 0.86 | NA |
| chr18a | 18 | 11098 | 0.30 | 1.01 | NA |
| chr20a | 20 | 2192 | 0.18 | 1.04 | NA |
| chr22a | 22 | 6156 | 0.07 | 0.51 | NA |
| chr25a | 25 | 3796 | 0.40 | 1.17 | NA |
| esc16a | 16 | 68 | 0.00 | 0.32 | 2.94 |
| esc16h | 16 | 996 | 0.00 | 0.10 | 0.00 |
| esc32a | 32 | 130 | 0.20 | 0.89 | NA |
| esc32e | 32 | 2 | 0.00 | 4.00 | 0.00 |
| esc32f | 32 | 2 | 0.00 | 6.00 | 0.00 |
| esc64a | 64 | 116 | 0.00 | 0.55 | NA |
| esc128 | 128 | 64 | 0.00 | 2.94 | NA |
| had12 | 12 | 1652 | 0.00 | 0.08 | 0.00 |
| had14 | 14 | 2724 | 0.00 | 0.09 | 0.07 |
| had16 | 16 | 3720 | 0.00 | 0.07 | 0.91 |
| had18 | 18 | 5358 | 0.00 | 0.06 | 1.12 |
| had20 | 20 | 6922 | 0.00 | 0.06 | 1.61 |
| kra30a | 30 | 88900 | 0.05 | 0.30 | NA |
| lipa20a | 20 | 3683 | 0.00 | 0.05 | NA |
| lipa20b | 20 | 27076 | 0.00 | 0.26 | NA |
| lipa30a | 30 | 13178 | 0.01 | 0.04 | 3.19 |
| lipa40a | 40 | 31538 | 0.01 | 0.03 | NA |
| lipa50a | 50 | 1210244 | 0.00 | 0.26 | 2.27 |
| lipa60a | 60 | 107218 | 0.00 | 0.02 | 2.09 |
| lipa70a | 70 | 169755 | 0.00 | 0.02 | 1.81 |
| lipa80a | 80 | 253195 | 0.00 | 0.02 | 1.65 |
| lipa90a | 90 | 360630 | 0.00 | 0.02 | 1.53 |
| nug12 | 12 | 578 | 0.01 | 0.17 | NA |
| nug14 | 14 | 1014 | 0.10 | 0.14 | NA |
| nug15 | 15 | 1150 | 0.00 | 0.17 | NA |
| nug16a | 16 | 1610 | 0.01 | 0.17 | NA |
| nug17 | 17 | 1732 | 0.01 | 0.17 | NA |
| nug18 | 18 | 1930 | 0.00 | 0.17 | 7.56 |
| nug20 | 20 | 2570 | 0.00 | 0.20 | NA |
| nug21 | 21 | 2438 | 0.00 | 0.24 | NA |
| nug22 | 22 | 3596 | 0.01 | 0.18 | NA |
| nug24 | 24 | 3488 | 0.00 | 0.20 | NA |
| nug25 | 25 | 3744 | 0.00 | 0.20 | NA |
| nug27 | 27 | 5234 | 0.01 | 0.23 | NA |
| nug28 | 28 | 5166 | 0.01 | 0.23 | NA |
| nug30 | 30 | 6124 | 0.00 | 0.20 | NA |
| rou12 | 12 | 235528 | 0.02 | 0.12 | NA |
| rou15 | 15 | 354210 | 0.03 | 0.17 | NA |
| rou20 | 20 | 725522 | 0.00 | 0.15 | NA |
| scr12 | 12 | 31410 | 0.00 | 0.28 | NA |
| scr15 | 15 | 51140 | 0.04 | 0.34 | NA |
| scr20 | 20 | 110030 | 0.00 | 0.75 | NA |
| sko42 | 42 | 15812 | 0.00 | 0.18 | NA |
| sko49 | 49 | 23386 | 0.01 | 0.17 | NA |
| sko56 | 56 | 34458 | 0.01 | 0.18 | NA |
| sko64 | 64 | 48498 | 0.01 | 0.16 | NA |
| sko72 | 72 | 66256 | 0.02 | 0.16 | NA |
| sko81 | 81 | 90998 | 0.01 | 0.14 | NA |
| sko90 | 90 | 115534 | 0.00 | 0.15 | NA |
| sko100a | 100 | 152002 | 0.01 | 0.14 | NA |
| ste36a | 36 | 9526 | 0.08 | 0.61 | NA |
| tai12a | 12 | 224416 | 0.01 | 0.18 | NA |
| tai15a | 15 | 388214 | 0.02 | 0.14 | NA |
| tai17a | 17 | 491812 | 0.02 | 0.15 | NA |
| tai20a | 20 | 703482 | 0.01 | 0.14 | NA |
| tai25a | 25 | 1167256 | 0.02 | 0.14 | 9.84 |
| tai30a | 30 | 1818146 | 0.03 | 0.16 | NA |
| tai35a | 35 | 2422002 | 0.03 | 0.15 | NA |
| tai40a | 40 | 3139370 | 0.02 | 0.14 | NA |
| tai50a | 50 | 4938796 | 0.03 | 0.14 | NA |
| tai60a | 60 | 7205962 | 0.02 | 0.14 | NA |
| tai64c | 64 | 1855928 | 0.00 | 0.17 | 5.31 |
| tai80a | 80 | 13499184 | 0.03 | 0.13 | 11.44 |
| tai100a | 100 | 21052466 | 0.02 | 0.12 | 11.02 |
| tai150b | 150 | 498896643 | 0.03 | 0.23 | NA |
| tai256c | 256 | 44759294 | 0.00 | 0.12 | NA |
| tho30 | 30 | 149936 | 0.04 | 0.21 | NA |
| tho40 | 40 | 240516 | 0.02 | 0.28 | NA |
| tho150 | 150 | 8133398 | 0.02 | 0.16 | NA |
| wil50 | 50 | 48816 | 0.00 | 0.10 | 7.65 |
| wil100 | 100 | 273038 | 0.00 | 0.08 | NA |

**Table 3** List of mQAP instances.

| Instance Name | Category | Size | Objective |
|---|---|---|---|
| KC10-2fl-1rl | real-like | 10 | 2 |
| KC10-2fl-2rl | real-like | 10 | 2 |
| KC10-2fl-3rl | real-like | 10 | 2 |
| KC10-2fl-4rl | real-like | 10 | 2 |
| KC10-2fl-5rl | real-like | 10 | 2 |
| KC20-2fl-1rl | real-like | 20 | 2 |
| KC20-2fl-2rl | real-like | 20 | 2 |
| KC20-2fl-3rl | real-like | 20 | 2 |
| KC20-2fl-4rl | real-like | 20 | 2 |
| KC20-2fl-5rl | real-like | 20 | 2 |
| KC50-2fl-1rl | real-like | 50 | 2 |
| KC50-2fl-2rl | real-like | 50 | 2 |
| KC50-2fl-3rl | real-like | 50 | 2 |
| KC10-2fl-1uni | uniform | 10 | 2 |
| KC10-2fl-2uni | uniform | 10 | 2 |
| KC10-2fl-3uni | uniform | 10 | 2 |
| KC20-2fl-1uni | uniform | 20 | 2 |
| KC20-2fl-2uni | uniform | 20 | 2 |
| KC20-2fl-3uni | uniform | 20 | 2 |
| KC50-2fl-1uni | uniform | 50 | 2 |
| KC50-2fl-2uni | uniform | 50 | 2 |
| KC50-2fl-3uni | uniform | 50 | 2 |

Indeed, we have not compared the time-consumption between different approaches because of the fact that GA approach ran on different platforms. Instead we compare the ultimate results to prove that the proposed method can found good solution regardless of size and time. There are many other studies that have been shown the time-consumption, but in order to compare the time we had to implement the different approaches and ran on the same machine. This might be one of the pitfalls of this paper and we leave this for future research.

We were assessed the performance of the proposed MOBFO using 22 mQAP instances (J. Knowles and Corne 2003). The experimental analyses environment as well as parameters initialization were the same as the proposed BFO algorithm according to Table 1.

Table 3 shows the list of 22 mQAP instances with their size and number of objectives. All of the instances are based on two objectives with different sizes (10-50) and categories (read-like and uniform). We have not assessed the proposed MOBFO on 3 dimensional problems and leave this part for future research.

We also evaluated the performance of the proposed MOBFO using three assessment metrics, average generational distance (GD), convergence measure (CM), and divergence measure (DM) (Mostaghim and Teich 2004; Okabe, Jin, and Sendhoff 2003). GD reports how far, in average, known Pareto front is from true Pareto front. In other words, GD metric evaluates the average distance between the nondominated set solutions (E) and optimum Pareto front (P*). GD can be defined as:

$$GD(E,P^*) = \frac{1}{|E|}\sum_{u\in E}\min\{dist(u,v)|v\in P^*\} \quad (7)$$

where, *dist*(*u, v*) is the Euclidean distance in objective space between the solution $u \in E$ and the nearest member in the optimum Pareto front ($v \in P^*$). Indeed, this metric measures how far the approximation front is from the optimum Pareto front. Note, the smaller value of GD represents a better performance.

CM evaluates average similarity of nondominated solutions and Pareto front solutions. CM can be defined as:

$$CM = \frac{\dfrac{1}{1+\left|\sum_{j=1}^{m}\sum_{k}^{n} pf(n)-nds(m,n)\right|}}{pf} \quad (8)$$

where, *pf* is Pareto optimum vector, *nds* is nondominated set solutions, *m* is size of nondominated set vector, and *n* is number of objectives. Note, the bigger the convergence measure, the better the result.

DM evaluates the spread distribution of vectors throughout the nondominated set solutions and can be defined as:

$$DM = d_{pf} + d_{nds} + \sqrt{\frac{1}{|nds|-1}\sum_{i=1}^{|nds|-1}\left(d_i-\bar{d}\right)^2} \quad (9)$$

where, $d_i$ is the Euclidean distance between nondominated set solutions, *i* is the number of nondominated set solutions, $d_{pf}$ and $d_{nds}$ are the Pareto optimum vector and nondominated set solutions. DM is expected to be better as the search area is wider, rather than single point, so the bigger the DM value, the better the result (Lim et al. 2014).

Consequently, we compared the results of these metrics (GD, CM, and DM) between proposed MOBFO and four well-known EA approaches: mGRASP/MH (Zinflou, Gagné, and Gravel 2013), Fuzzy PSO (M. Zhao et al. 2008), NSGAII (Deb et al. 2002), and original MOBFO (Niu et al. 2013). Table 4 shows the performance of assessment metrics obtained by each algorithm for the 22 mQAP instances.

The first column is the instance name, the second column is the evaluation metric name, and the rest of the columns indicate the results for each algorithm. Based on the results, the proposed MOBFO performed better than other algorithms in terms of the average GD, CM, and DM values. Other methods have either large GD value or small CM and DM values.

**Table 4** Average Generational Distance (GD), Convergence Measure (CM), and Divergence Measure (DM) values for mGRASP/MH, Fuzzy PSO, NSGAII, original MOBFO, and proposed MOBFO. First and second columns are the name of instances and metrics, and the rest of the columns compare the performance of algorithms. NA indicates not available.

| Instance Name | Metric | mGRASP/MH | Fuzzy PSO | NSGAII | Original MOBFO | Proposed MOBFO |
|---|---|---|---|---|---|---|
| KC10-2fl-1rl | GD | 6.0364e+04 | 0 | 0 | 3.6304e+04 | 0 |
| | CM | 0.0173 | 0.2242 | 1 | 0.0865 | 1 |
| | DM | 0.2069 | 0.2241 | 1 | 0.3103 | 1 |
| KC10-2fl-2rl | GD | 7.7505e+04 | 0 | 0 | 0 | 0 |
| | CM | 0.0667 | 0.2667 | 1 | 0.6668 | 1 |
| | DM | 0.3333 | 0.2667 | 1 | 0.6667 | 1 |
| KC10-2fl-3rl | GD | 6.5790e+04 | 0 | 0 | 2.0828e+04 | 0 |
| | CM | 1.0486e-05 | 0.1455 | 1 | 0.1092 | 1 |
| | DM | 0.0909 | 0.1455 | 1 | 0.2545 | 1 |
| KC10-2fl-4rl | GD | 1.0145e+04 | 0 | 0 | 2.2742e+03 | 0 |
| | CM | 0.3435 | 0.2283 | 1 | 0.2492 | 1 |
| | DM | 0.4340 | 0.2264 | 1 | 0.3208 | 1 |
| KC10-2fl-5rl | GD | 3.7627e+04 | NA | 0 | 0 7.8350e+04 | 0 |
| | CM | 0.2858 | NA | 1 | 0.2042 | 1 |
| | DM | 0.4490 | NA | 1 | 0.3061 | 1 |
| KC20-2fl-1rl | GD | 1.0004e+06 | NA | 8.6505e+04 | 3.3506e+05 | 1.4131e+03 |
| | CM | 1.1602e-05 | NA | 7.8179e-05 | 4.0738e-05 | 0.9140 |
| | DM | 0.1290 | NA | 0.4194 | 0.1398 | 0.9570 |
| KC20-2fl-2rl | GD | 5.2403e+06 | NA | 2.5075e+06 | 4.8649e+06 | 1.4435e+06 |
| | CM | 1.1227e-05 | NA | 1.0746e-04 | 1.8974e-05 | 0.5366 |
| | DM | 0.0909 | NA | 0.2455 | 0.0818 | 0.7909 |
| KC20-2fl-3rl | GD | 1.0763e+05 | NA | 4.1562e+04 | 4.0651e+06 | 7.9226e+03 |
| | CM | 2.0048e-05 | NA | 1.6249e-04 | 1.0101e-05 | 0.2711 |
| | DM | 0.0714 | NA | 0.3878 | 0.0408 | 0.5459 |
| KC20-2fl-4rl | GD | 9.4466e+05 | NA | 3.1785e+05 | 1.4372e+06 | 9.7568e+04 |
| | CM | 7.8523e-06 | NA | 0.0137 | 6.0072e-06 | 0.2467 |
| | DM | 0.1644 | NA | 0.4658 | 0.1096 | 0.6438 |
| KC20-2fl-5rl | GD | 7.8927e+05 | NA | 2.3552e+05 | 1.0829e+06 | 8.1897e+04 |
| | CM | 4.2539e-06 | NA | 3.3240e-05 | 1.9945e-06 | 0.0807 |
| | DM | 0.1129 | NA | 0.3952 | 0.0645 | 0.5000 |
| KC50-2fl-1rl | GD | NA | NA | NA | 1.8848e+06 | 2.4066e+04 |
| | CM | NA | NA | NA | 2.7198e-07 | 2.4554e-04 |
| | DM | NA | NA | NA | 0.0139 | 0.5278 |
| KC50-2fl-2rl | GD | NA | NA | NA | 2.1147e+06 | 3.5228e+04 |
| | CM | NA | NA | NA | 6.3955e-08 | 1.6289e-04 |
| | DM | NA | NA | NA | 0.0100 | 0.5116 |
| KC50-2fl-3rl | GD | NA | NA | NA | 1.2676e+06 | 2.7726e+04 |
| | CM | NA | NA | NA | 9.5983e-07 | 1.7589e-04 |
| | DM | NA | NA | NA | 0.0156 | 0.4630 |
| KC10-2fl-1uni | GD | 2.3550e+03 | 0 | 0 | 6.2979e+03 | 0 |
| | CM | 0.1554 | 1 | 1 | 0.0015 | 1 |
| | DM | 0.4615 | 1 | 1 | 0.4615 | 1 |
| KC10-2fl-2uni | GD | 8.5809e+03 | 0 | 0 | 0 | 0 |
| | CM | 1.2117e-04 | 1 | 1 | 1 | 1 |
| | DM | 1 | 1 | 1 | 1 | 1 |
| KC10-2fl-3uni | GD | 462.0308 | 0 | 0 | 427.8507 | 0 |
| | CM | 0.0488 | 1 | 1 | 0.0104 | 1 |
| | DM | 0.2077 | 1 | 1 | 0.1923 | 1 |
| KC20-2fl-1uni | GD | 1.4341e+04 | NA | 3.0844e+03 | 1.1162e+04 | 699.2820 |
| | CM | 9.6647e-04 | NA | 0.0026 | 8.2174e-04 | 0.5472 |
| | DM | 0.1818 | NA | 0.4727 | 0.1273 | 0.7455 |
| KC20-2fl-2uni | GD | 5.3530e+05 | NA | 5.2399e+05 | 8.5236e+05 | 3.4735e+05 |
| | CM | 1.6519e-04 | NA | 7.8772e-04 | 7.4803e-06 | 0.0059 |
| | DM | 0.3750 | NA | 0.5000 | 0.1250 | 0.7500 |
| KC20-2fl-3uni | GD | 2.1538e+03 | NA | 1.1816e+03 | 5.3478e+03 | 155.1058 |
| | CM | 1.8263e-07 | NA | 1.8208e-07 | 1.7929e-07 | 1.8413e-07 |
| | DM | 0.0051 | NA | 0.0051 | 0.0102 | 0.0051 |
| KC50-2fl-1uni | GD | NA | NA | NA | 7.5229e+04 | 2.3338e+03 |
| | CM | NA | NA | NA | 1.0256e-05 | 7.7719e-04 |
| | DM | NA | NA | NA | 0.0471 | 0.5294 |
| KC50-2fl-2uni | GD | NA | NA | NA | 2.2387e+05 | 6.0161e+03 |
| | CM | NA | NA | NA | 1.5426e-06 | 2.9983e-04 |
| | DM | NA | NA | NA | 0.0588 | 0.4118 |
| KC50-2fl-3uni | GD | NA | NA | NA | 2.3057e+04 | 417.7430 |
| | CM | NA | NA | NA | 1.0802e-04 | 0.0035 |
| | DM | NA | NA | NA | 0.0378 | 0.5689 |

Additionally, we illustrated the graphical performance comparison of these five algorithms including optimum Pareto front in Figure 1 for all of the instances listed in Table 4. Although, the proposed MOBFO outperforms NSGAII, Fuzzy PSO, mGRASP/MH, and original MOBFO, but in very rare cases the approximation sets found by the proposed MOBFO are very close to Pareto optimum solutions. For the future direction, this can be resolved by hyperparameter optimization for the constant values in Table 1 using nested cross-validation (Parvandeh et al. 2020).

## 6 Conclusion

In this paper, we proposed modified BFO and MOBFO algorithms for the solution of QAP and mQAP instances. The proposed BFO algorithm updates the solution using the swap mutation method in chemotactic part and improves the final solution using local tabu search algorithm to reach local optimum. And, the proposed MOBFO attempts to find a nondominated set solutions to be as close as possible to Pareto front solutions. MOBFO determines the nondominated set solutions using fast nondominated sort method and updates solutions set using swap mutation and uniform like crossover operation (ULX). Anditionally, the final nondominated set solutions improves using multi-objective version of local tabu search to reach the local optimum.

We assessed the proposed BFO and MOBFO algorithms on QAP and 22 mQAP instances from the QAPLIB. Computational results showed that BFO and MOBFO performed well on these instances. We compared both algorithms with the number of state-of-the-art algorithms as well as the best-known solutions from QAPLIB. The comparison showed that our results are very close to the best-known solutions and outperforms the state-of-the-art algorithms.

**Figure 1** Comparison results of five algorithms, proposed MOBFO, original BFO, mGRASP/MH, Fuzzy PSO, and NSGA2. In this figure KC10/KC20 related to problem size, and 2fl is related to number of objective functions. Also, rl/uni related to type of the problem whether is real (rl) or uniform (uni).

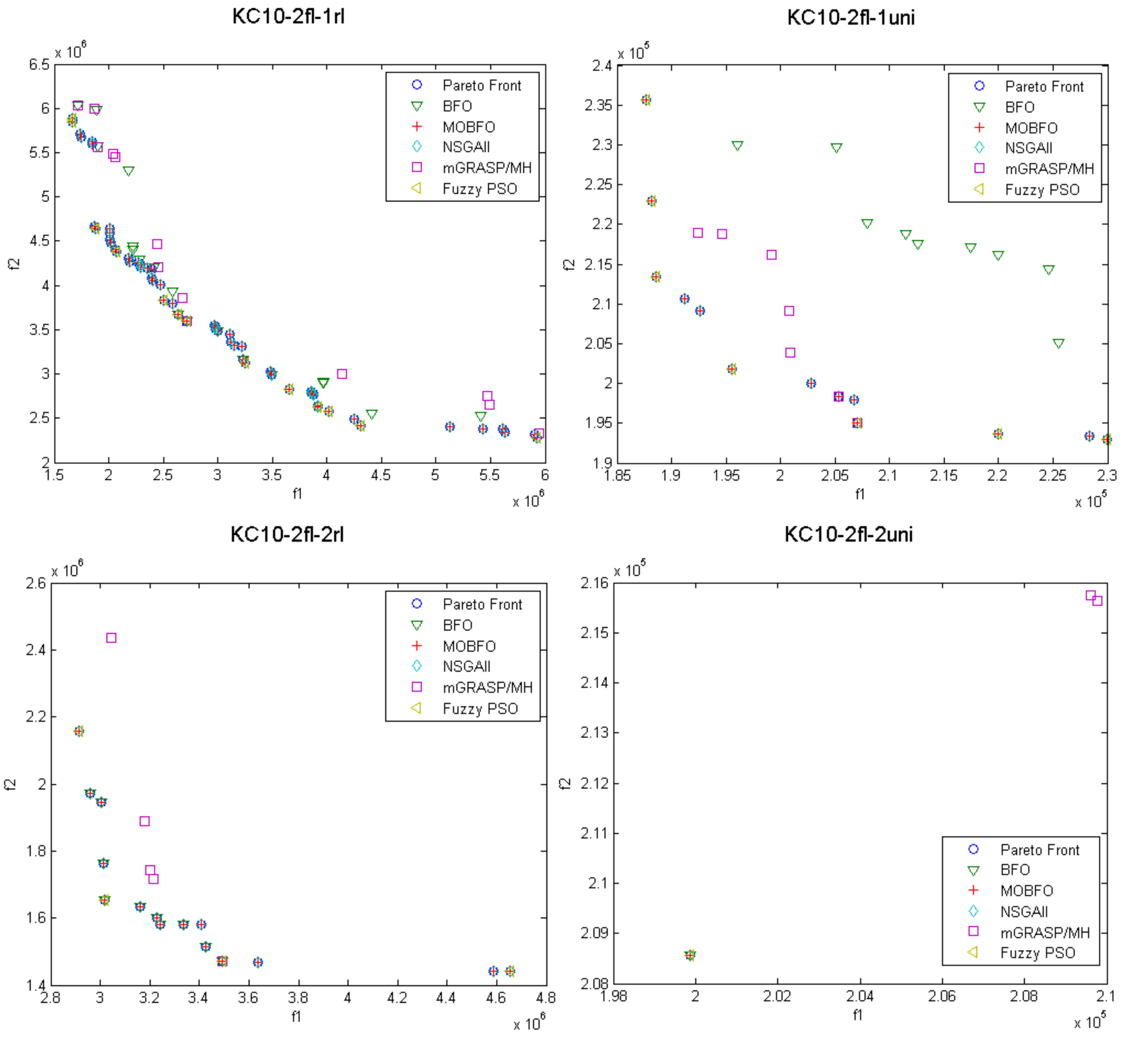

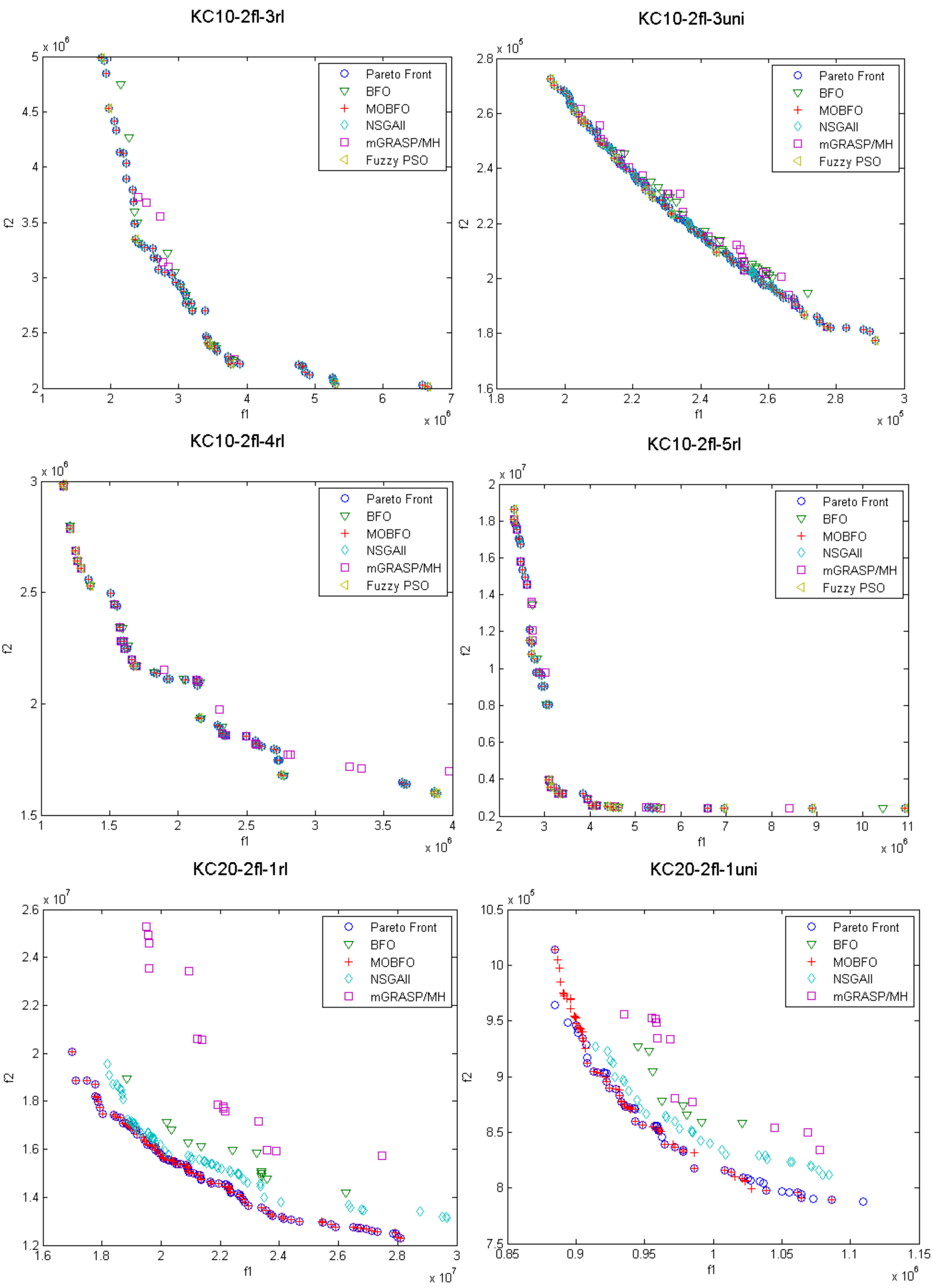
KC10-2fl-3rl
x 10^6
5
4.5
4
3.5
3
2.5
2
f2
1 2 3 4 5 6 7
f1
x 10^6
Pareto Front
BFO
MOBFO
NSGAII
mGRASP/MH
Fuzzy PSO
KC10-2fl-3uni
x 10^5
2.8
2.6
2.4
2.2
2
1.8
1.6
f2
1.8 2 2.2 2.4 2.6 2.8 3
f1
x 10^5
Pareto Front
BFO
MOBFO
NSGAII
mGRASP/MH
Fuzzy PSO
KC10-2fl-4rl
x 10^6
3
2.5
2
1.5
f2
1 1.5 2 2.5 3 3.5 4
f1
x 10^6
Pareto Front
BFO
MOBFO
NSGAII
mGRASP/MH
Fuzzy PSO
KC10-2fl-5rl
x 10^7
2
1.8
1.6
1.4
1.2
1
0.8
0.6
0.4
0.2
f2
2 3 4 5 6 7 8 9 10 11
f1
x 10^6
Pareto Front
BFO
MOBFO
NSGAII
mGRASP/MH
Fuzzy PSO
KC20-2fl-1rl
x 10^7
2.6
2.4
2.2
2
1.8
1.6
1.4
1.2
f2
1.6 1.8 2 2.2 2.4 2.6 2.8 3
f1
x 10^7
Pareto Front
BFO
MOBFO
NSGAII
mGRASP/MH
KC20-2fl-1uni
x 10^5
10.5
10
9.5
9
8.5
8
7.5
f2
0.85 0.9 0.95 1 1.05 1.1 1.15
f1
x 10^6
Pareto Front
BFO
MOBFO
NSGAII
mGRASP/MH

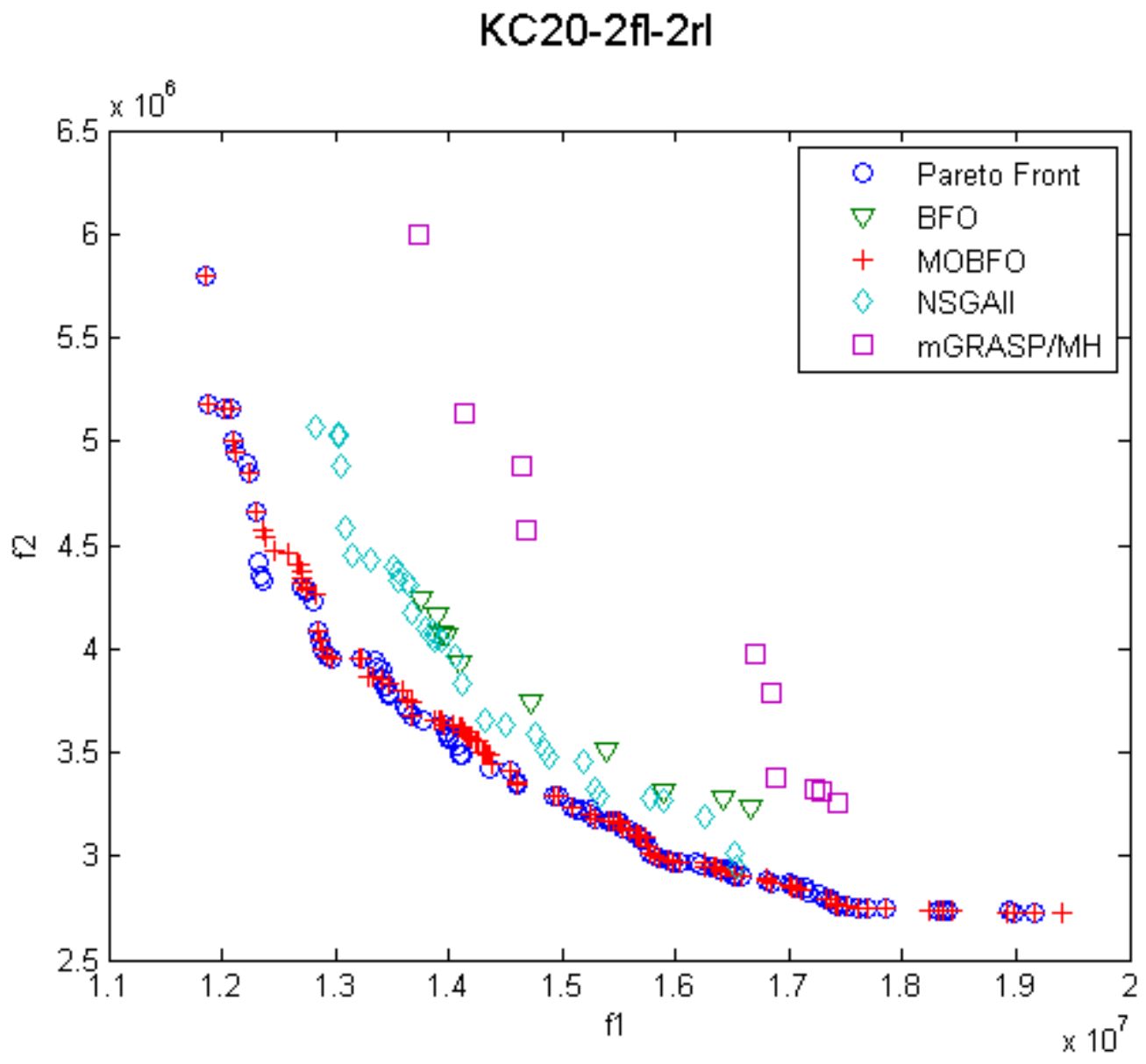
KC20-2fl-2rl
x 10^6
f2
f1
x 10^7
Pareto Front
BFO
MOBFO
NSGAII
mGRASP/MH

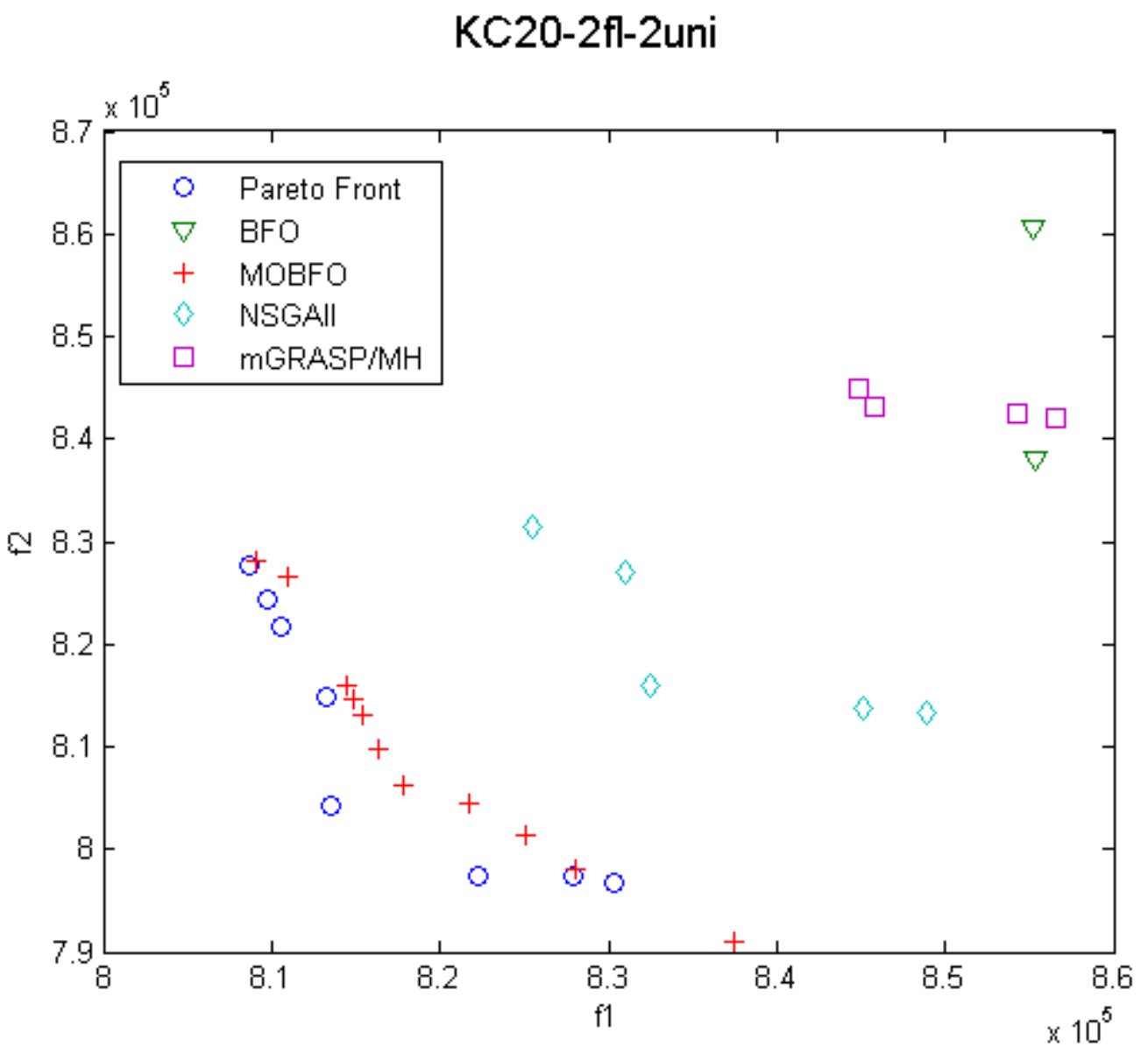
KC20-2fl-2uni
x 10^5
f2
f1
x 10^5
Pareto Front
BFO
MOBFO
NSGAII
mGRASP/MH

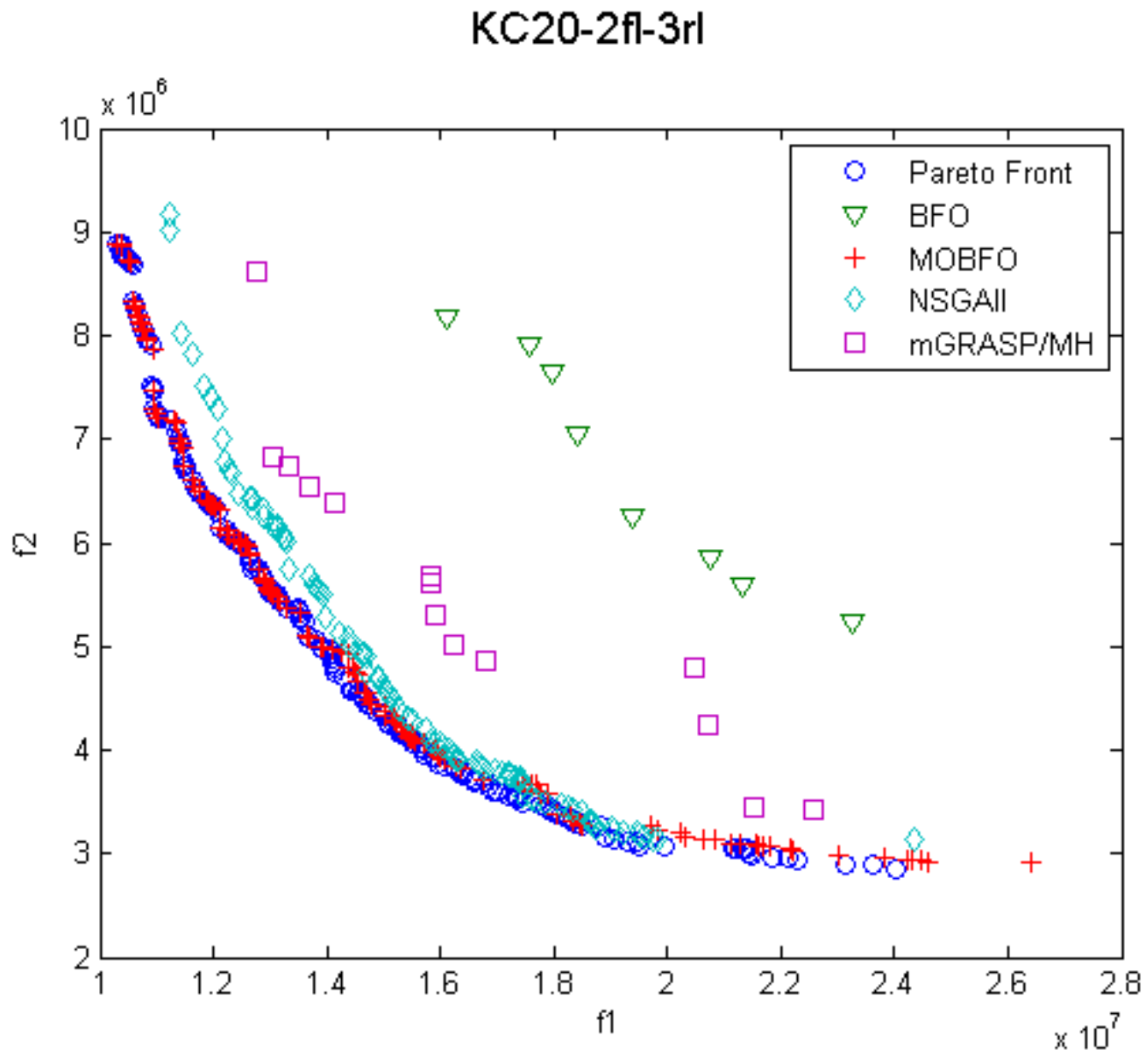
KC20-2fl-3rl
x 10^6
f2
f1
x 10^7
Pareto Front
BFO
MOBFO
NSGAII
mGRASP/MH

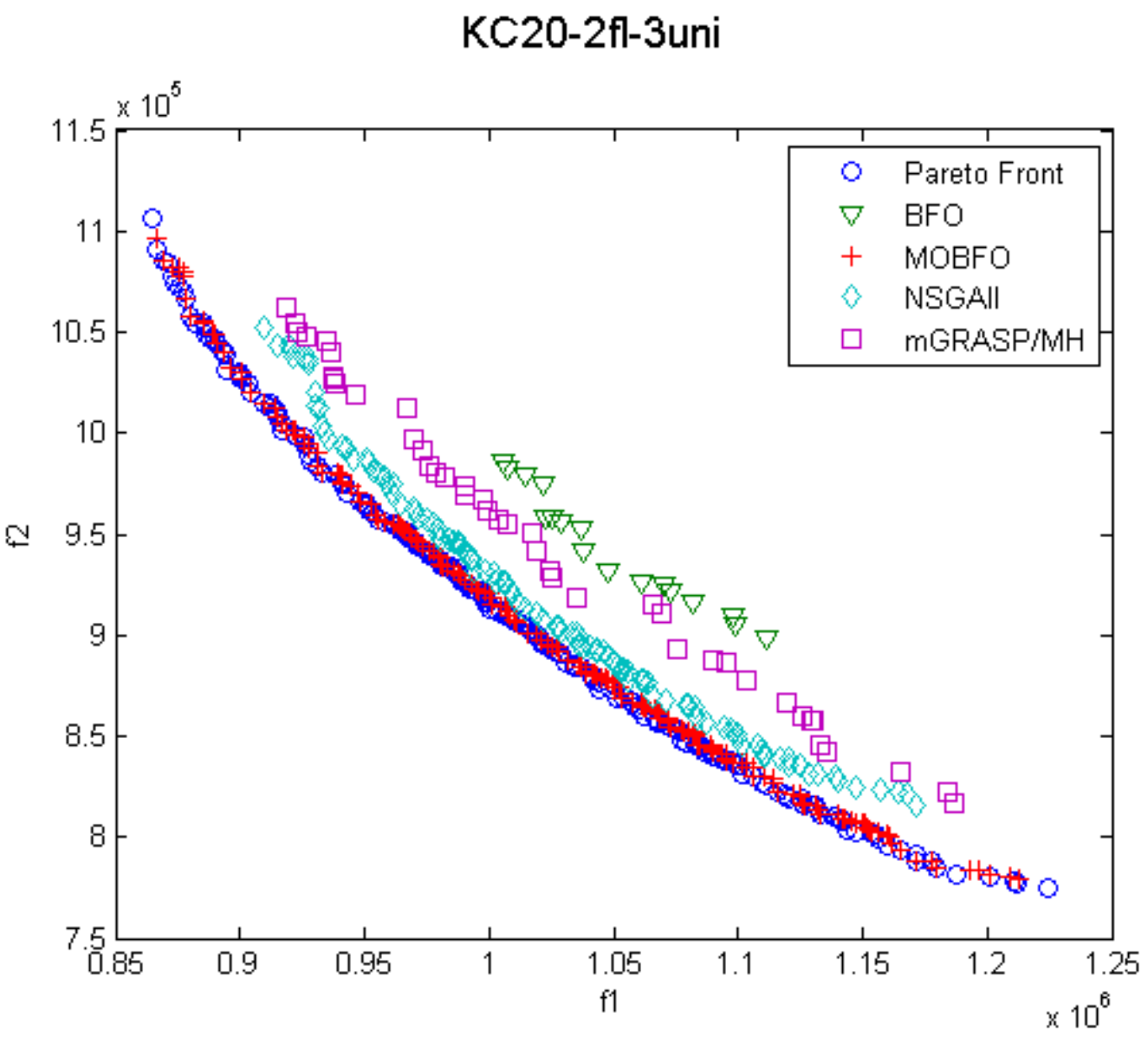
KC20-2fl-3uni
x 10^5
f2
f1
x 10^6
Pareto Front
BFO
MOBFO
NSGAII
mGRASP/MH

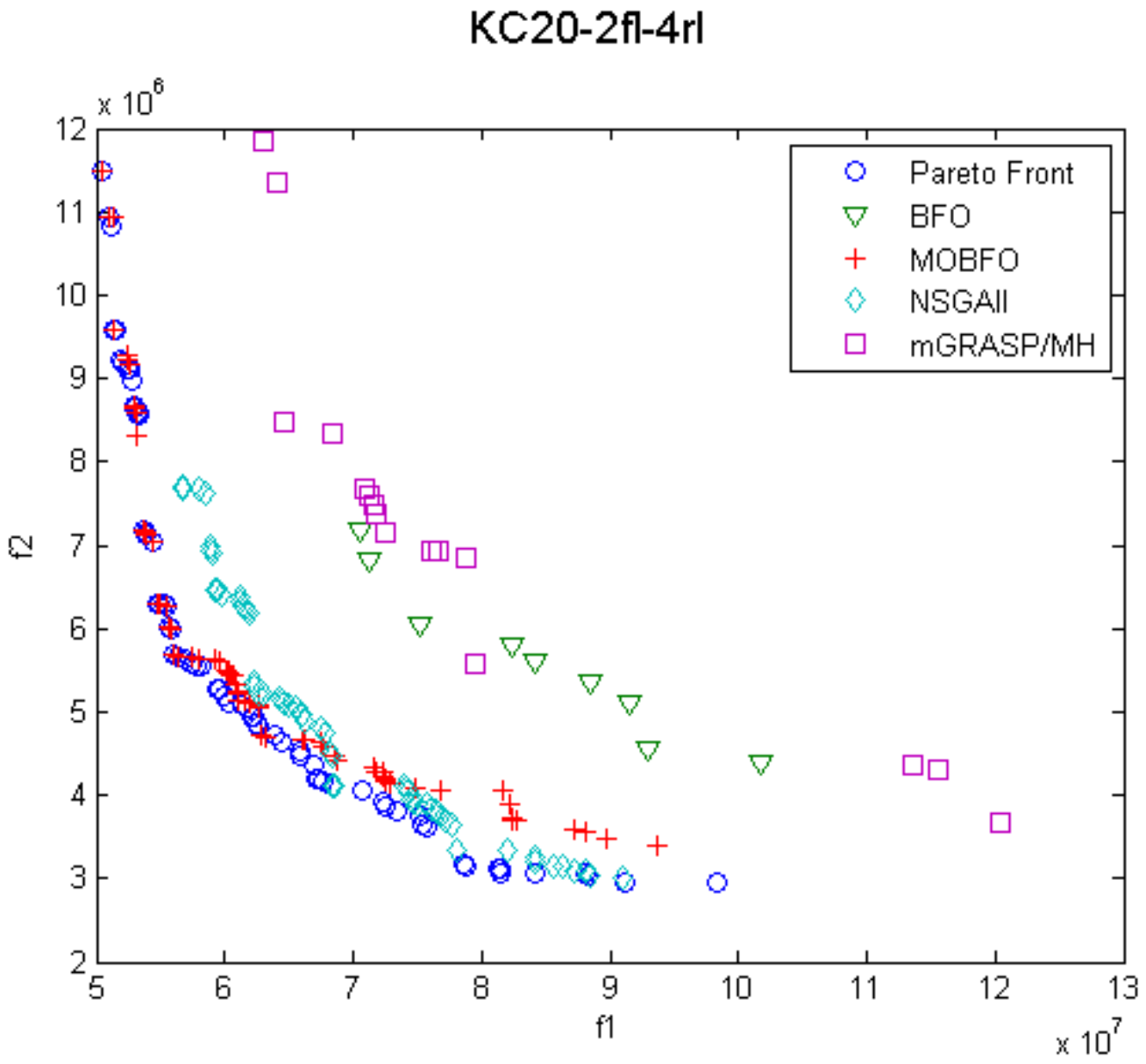
KC20-2fl-4rl
x 10^6
f2
f1
x 10^7
Pareto Front
BFO
MOBFO
NSGAII
mGRASP/MH

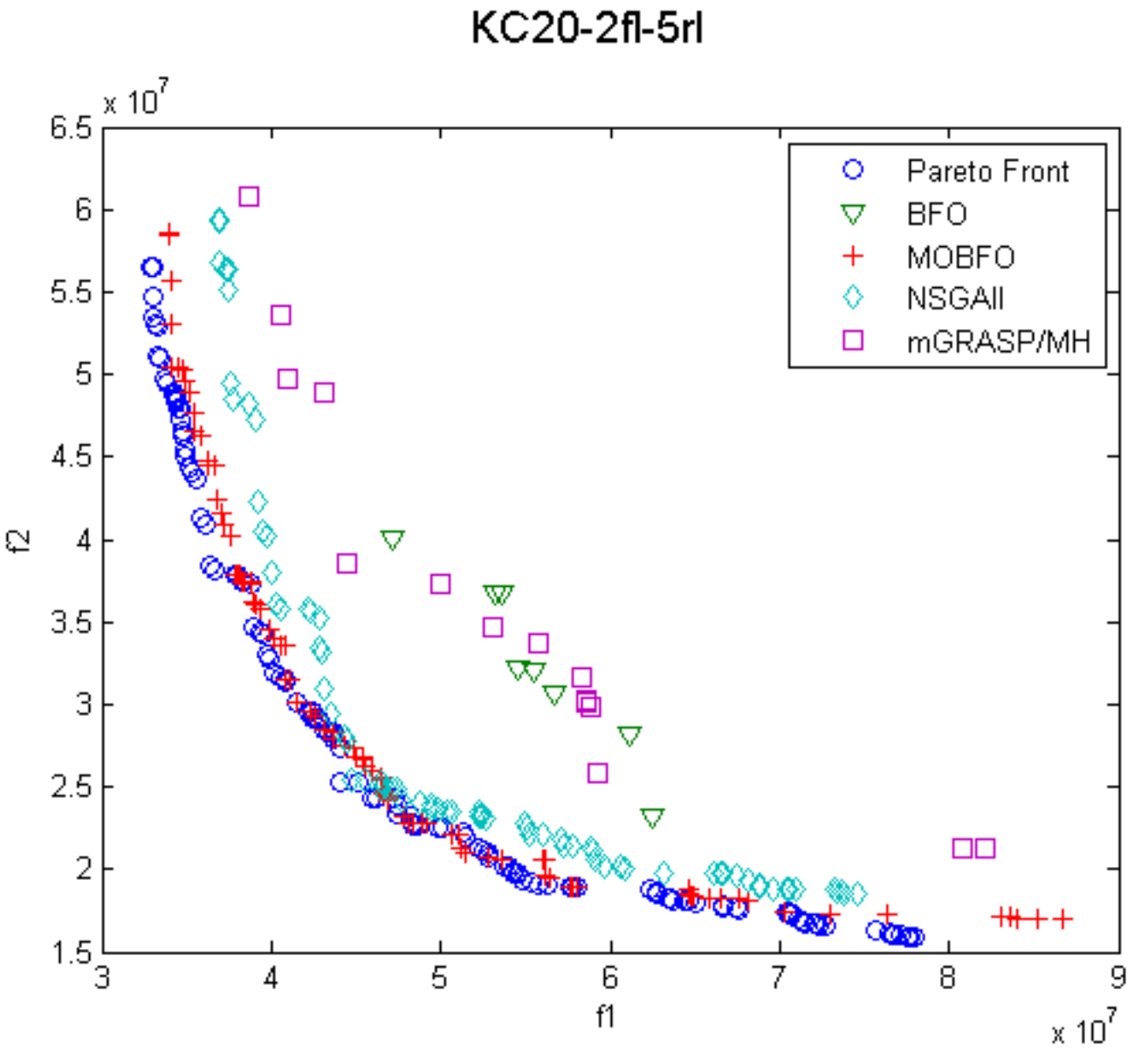
KC20-2fl-5rl
x 10^7
f2
f1
x 10^7
Pareto Front
BFO
MOBFO
NSGAII
mGRASP/MH

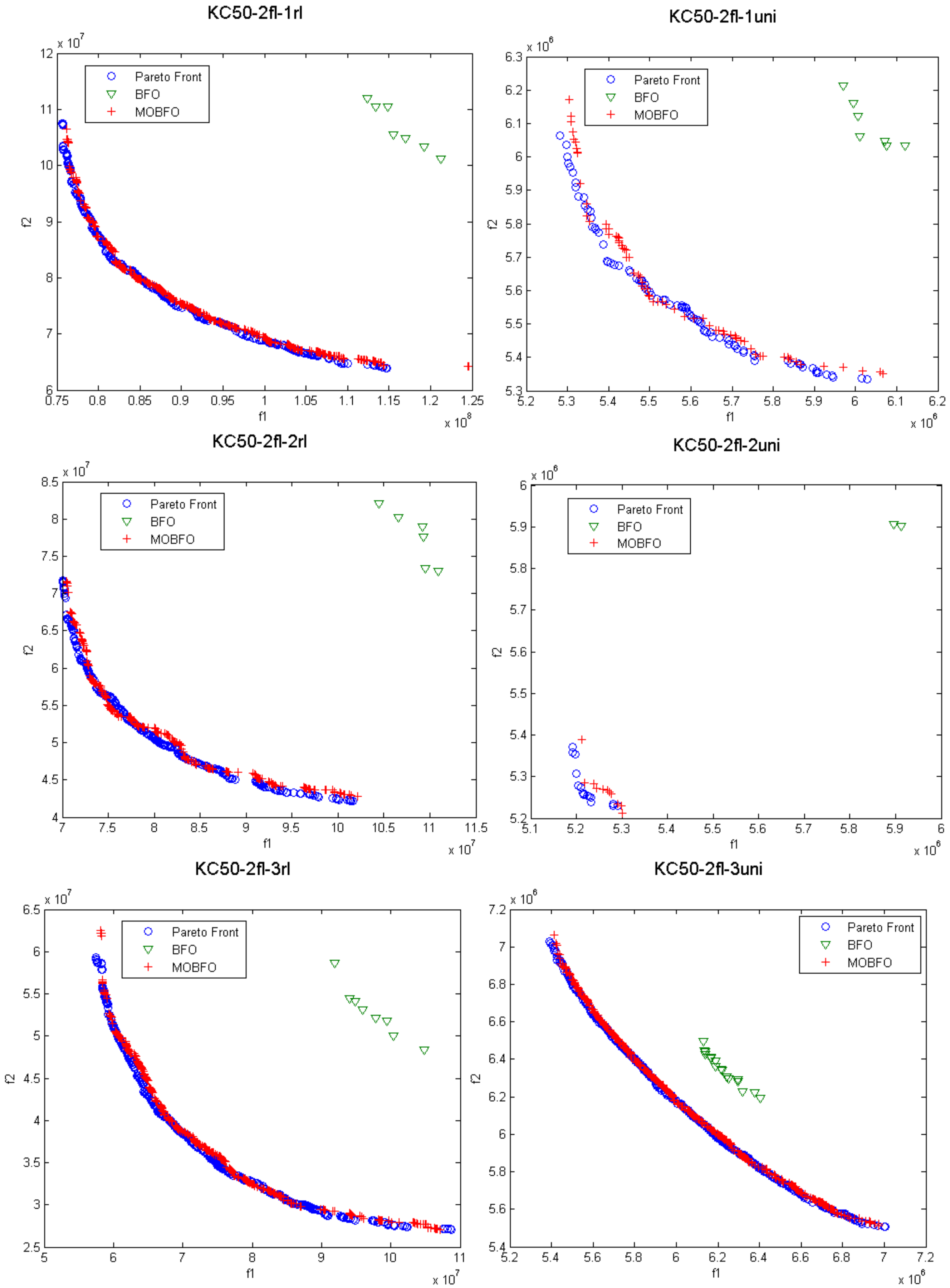
KC50-2fl-1rl
Pareto Front
BFO
MOBFO
f1
f2
x 10^8
x 10^7
KC50-2fl-1uni
Pareto Front
BFO
MOBFO
f1
f2
x 10^6
x 10^6
KC50-2fl-2rl
Pareto Front
BFO
MOBFO
f1
f2
x 10^7
x 10^7
KC50-2fl-2uni
Pareto Front
BFO
MOBFO
f1
f2
x 10^6
x 10^6
KC50-2fl-3rl
Pareto Front
BFO
MOBFO
f1
f2
x 10^7
x 10^7
KC50-2fl-3uni
Pareto Front
BFO
MOBFO
f1
f2
x 10^6
x 10^6